\documentclass[journal,web]{ieeecolor}
\usepackage{generic}
\usepackage{cite}
\usepackage{amsmath,amssymb,amsfonts}
\usepackage{algorithmic}
\usepackage{multirow}
\usepackage{subcaption}
\usepackage{graphicx}
\usepackage{algorithm,algorithmic}
\usepackage{hyperref}
\hypersetup{hidelinks=true}
\usepackage{textcomp}
\def\BibTeX{{\rm B\kern-.05em{\sc i\kern-.025em b}\kern-.08em
    T\kern-.1667em\lower.7ex\hbox{E}\kern-.125emX}}
\begin{document}
\title{Exploring the Impact of Skin Color on Skin Lesion Segmentation}
\author{Kuniko Paxton, Medina Kapo, Amila Akagić, Koorosh Aslansefat, Dhavalkumar Thakker and Yiannis Papadopoulos
\thanks{1 March 2026}
\thanks{Kuniko Paxton is with the School of Digital and Physical Sciences at the University of Hull, Cottingham Road, Hull, HU6 7RX, United Kingdom (e-mail: k.azuma-2021@hull.ac.uk).}
\thanks{Medina Kapo is with the Faculty of Electrical Engineering at the University of Sarajevo, Zmaja od Bosne bb, Sarajevo, 71000, Bosnia and Herzegovina (e-mail: mkapo2@etf.unsa.ba).}
\thanks{Amila Akagić is with the Faculty of Electrical Engineering at the University of Sarajevo, Zmaja od Bosne bb, Sarajevo, 71000, Bosnia and Herzegovina (e-mail: aakagic@etf.unsa.ba).}
\thanks{Koorosh Aslansefat is with the School of Digital and Physical Sciences at the University of Hull, Cottingham Road, Hull, HU6 7RX, United Kingdom (e-mail: K.Aslansefat@hull.ac.uk).}
\thanks{Dhavalkumar Thakker is with the School of Digital and Physical Sciences at the University of Hull, Cottingham Road, Hull, HU6 7RX, United Kingdom (e-mail: D.Thakker@hull.ac.uk).}
\thanks{Yiannis Papadopoulos is with the School of Digital and Physical Sciences at the University of Hull, Cottingham Road, Hull, HU6 7RX, United Kingdom (e-mail: Y.I.Papadopoulos@hull.ac.uk).}}

\maketitle

\begin{abstract}
Skin cancer, particularly melanoma, remains a major cause of morbidity and mortality, making early detection critical. AI-driven dermatology systems often rely on skin lesion segmentation as a preprocessing step to delineate the lesion from surrounding skin and support downstream analysis. While fairness concerns regarding skin tone have been widely studied for
lesion classification, the influence of skin tone on the segmentation stage remains under-quantified and is frequently assessed using coarse, discrete skin-tone categories. In this work, we evaluate three strong segmentation architectures (UNet, DeepLabV3 with a ResNet-50 backbone, and DINOv2) on two public dermoscopic datasets (HAM10000 and ISIC2017) and introduce a continuous pigment/contrast analysis that treats pixel-wise ITA values as distributions. Using Wasserstein distances between "within-image distributions" for skin-only, lesion-only, and whole-image regions, we quantify lesion–skin contrast and relate it to segmentation performance across multiple metrics. Within the range represented in these datasets, global skin tone
metrics (Fitzpatrick grouping or mean ITA) show weak association with segmentation quality. In contrast,
low lesion–skin contrast is consistently associated with larger segmentation errors in models, indicating that boundary ambiguity and low contrast are key drivers of failure. These findings suggest that fairness improvements in dermoscopic segmentation should prioritize robust handling of low-contrast lesions, and the distribution-based pigment measures provide a more informative audit signal than discrete skin-tone categories.
\end{abstract}


\section{Introduction}
\label{sec:introduction}
Melanoma is one of the deadliest forms of skin cancer, with an estimated 325,000 new cases and 57,000 deaths worldwide in 2020 (WHO). Early detection and treatment dramatically improve survival rates, driving the development of advanced AI tools for diagnosing skin lesions. They have now achieved expert-level accuracy in distinguishing malignant from benign lesions. Many high-performing AI-driven lesion diagnosis models employ pipelines that include hair removal \cite{kasmi2023sharprazor,rizzi2020skin} and skin lesion segmentation as a preprocessing step to focus analysis on the lesion area within the image. In such segmentation-based pipelines, lesions are first automatically outlined and then cropped or highlighted. This enables subsequent classifiers to minimize the surrounding skin area and focus primarily on examining the lesion itself. Research has demonstrated that combining segmentation and classification helps models focus on relevant image regions, thereby enhancing diagnostic performance \cite{lopez2022interpretable}. 

Despite segmentation playing a central role in skin lesion diagnosis, it remains sensitive to variations in skin tone, which is a well-known challenge in this field \cite{pakzad2022circle, kinyanjui2019estimating, rezk2022improving, montoya2024towards}. By isolating lesions, segmentation excludes most of the surrounding skin (which conveys the patient's pigmentation), potentially reducing the influence of skin tone on the classifier's decision. Intuitively, perfect segmentation should render the classifier indifferent to skin tone, allowing it to focus solely on lesion features. Despite segmentation's intuitive benefits for fairness, the impact of skin color on the segmentation stage remains unresolved. When segmentation algorithms become more challenging for specific skin tones (e.g., very dark or very light skin), bias may arise in earlier stages of the diagnostic process. Conversely, if segmentation performance remains consistent across skin tones, it suggests that this preprocessing step is not a source of fairness weakness.

Therefore, this manuscript investigates the impact of skin tone on skin lesion segmentation. Semantic segmentation provides pixel-level predictions; therefore, the methods used in traditional skin classification studies, which divide skin color into a few categories for comparison, are unable to fully capture its impact. Hence, this research investigates the influence of skin tone using metrics that preserve individual skin tone nuances and unevenness, rather than relying on broad skin color groups in the analysis. The widely used CNN-based segmentation models UNET~\cite{ronneberger2015u} and DeepLabV3 \cite{chen2017rethinking}, alongside the relatively newer Vision Transformers-based model DINOv2~\cite{oquab2023dinov2}, are employed to cover the diverse models. The Human Against Machine with 10000 training images (HAM) dataset \cite{tschandl2018ham10000, tschandl2020human} and the 2017 International Skin Image Collaborative Research (ISIC) dataset \cite{codella2018skin} were validated. These are among the most widely used datasets for skin lesion classification and include publicly available ground truth annotations for segmentation. Our systematic analysis of skin lesion segmentation promotes fairness across diverse skin tones, encouraging the design of improved segmentation methods and contributing to safer, fairer models.

\section{Related Work}
To our knowledge, research addressing skin tone bias in dermatological lesion segmentation remains limited. Therefore, we investigate previous research from two perspectives: first, the impact of segmentation on skin classifiers \ref{sec:related1}; second, the issue of fairness in tasks related to skin lesions.

\subsection{Impact of Segmentation on Skin Classifiers}\label{sec:related1}
Lesion segmentation is an essential element in skin lesion classification, and numerous studies have been proposed to date. This subsection focuses exclusively on existing research on skin lesion segmentation that analyzes not only the performance of the segmentation methods themselves but also their relationships and interactions with classifiers. \cite{gamage2024melanoma} have successfully incorporated U2-Net into their pipeline, significantly improving the accuracy, sensitivity, and specificity. YoTransViT \cite{saha2024yotransvit} achieves improved classification performance by integrating data augmentation and segmentation. \cite{singh2022skinet} proposed a pipeline that incorporates segmentation and classification, aiming to enhance model reliability and classification accuracy. This pipeline considered the impact of uncertainty inherent in segmentation on classification results. This is based on the awareness that uncertainty of its segmentation may degrade subsequent classification performance. \cite{mahbod2020effects} empirically demonstrated that the post-segmentation processing method had a significant impact on classification accuracy. \cite{minhas2020accurate} suggests that with sufficiently accurate segmentation, other preprocessing steps can be omitted. Additionally, literature reviews of skin lesion classification have reported that DL models achieve high classification performance with appropriate segmentation \cite{adegun2021deep,khan2023identifying,zhang2023recent}. These studies clearly highlight the significant impact of lesion segmentation on skin lesion classification, which is a central theme of our research in this paper. We advance this strand of work by presenting the results of the analysis from a novel perspective, which used statistical distance to quantify the contrast between skin and lesion color.

\subsection{Issue of Fairness in Tasks Related to Skin Lesions}\label{sec:related2}
It is widely recognized that skin tone diversity is insufficiently represented in dermatological lesion datasets. This issue has been highlighted in literature reviews by \cite{guo2022bias} and \cite{wen2022characteristics}, as well as in dataset studies by \cite{pope2024skin}. These studies emphasize the importance of considering diverse skin types. Consequently, skin lesion classification tasks employ diverse approaches, including image generation \cite{pakzad2022circle,rezk2022improving,abhari2023mitigating}, federated learning \cite{xu2022achieving,fan2021fairness}, model structure reduction \cite{kong2024achieving,wu2022fairprune,paxton2026enhancing}, and disentangled representation learning \cite{aayushman2024fair,du2022fairdisco}. However, these approaches did not account for the fact that segmentation can alter the skin region.  \cite{chiu2024achieve} proposed a method to reduce skin color effects through segmentation. However, it did not address the possibility that the segmentation model itself might contain bias. Current bias reduction techniques demonstrate some effectiveness, yet they do not fundamentally resolve the bias issue. Despite the fact that further research is required \cite{bissoto2020debiasing} on skin color bias in segmentation models, there is little comprehensive research. To the best of our knowledge, the only other study to evaluate the existence of bias in lesion segmentation models is that of \cite{benvcevic2024understanding}. This study was the first research to assess fairness in skin lesion segmentation comprehensively, and it provided insights for future research in this field. However, the results of this study are limited for the following reasons. First, the three methods used to evaluate skin color are the Fitzpatrick skin color scale, Individual Typology Angle (ITA), and manual grouping. All those methods categorize skin color into discrete groups and ignore skin color gradations or color differences within images. Second, the study only used UNET-based models and did not investigate models based on different architectures, such as DeepLabV3 and Vision Transformer-based DINOv2. Therefore, it is unclear whether their results would hold for models other than UNET. Our study aims to provide a more precise analysis of the relationship between skin color and segmentation. Our analysis provides deeper insights by capturing the statistical distribution of skin color in a continuum and comparing the impact of bias across different segmentation models.

\subsection{Research Questions}
Although lesion segmentation is widely used in the dermatological pipeline, the relationship between semantic segmentation accuracy and skin color has not yet been quantitatively characterized at the pixel level. To address this gap, we pose the following research questions.
\begin{enumerate}
    \item Does skin lesion segmentation performance differ across patient skin tones, and if so, by how much?
    \item How does lesion-skin pigment contrast affect segmentation accuracy?
    \item How do segmentation error rates vary as a function of lesion-skin pigment contrast?
\end{enumerate}
\subsection{Main Contribution}
This study provides a quantitative analysis of how skin tone and lesion-skin contrast relate to lesion segmentation performance. The main methodological contribution is a distribution-based skin tone representation that preserves within-image gradations and unevenness, enabling contrast to be quantified without reducing tone to a small number of discrete categories.

\section{Methods}
\begin{figure}
    \centering
    \includegraphics[width=1\linewidth]{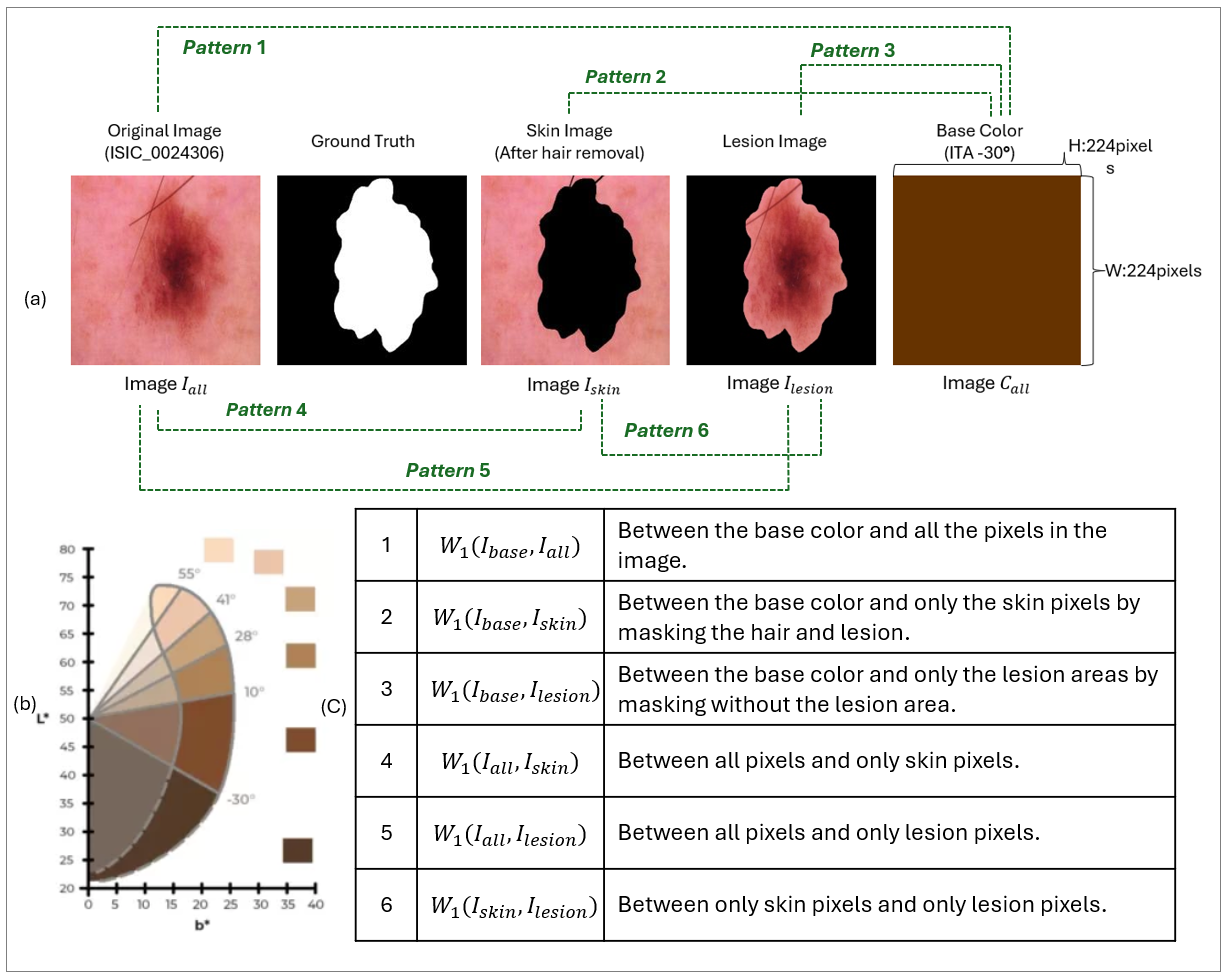}
    \caption{Research Analysis Overview: After semantic segmentation, the skin tone is measured by green colored six comparison patterns in (a). From left to right, the original image, the ground truth annotation, and the image in which the annotations have been used to remove the hair and extract only the skin-colored pixels. The fourth image shows annotations used to measure only the lesion's color, and then the base color image. The image (b) reproduced from \cite{medelink} illustrates the ITA that identified pixel-wise color links to skin. (c) is a detailed description of six comparison patterns.}
    \label{fig:main}
\end{figure}

Figure \ref{fig:main} summarizes the analysis pipeline. After semantic segmentation, we compute ITA at the pixel level and compare six color-distribution patterns (Figure \ref{fig:main}(c)) to disentangle overall tone, skin-only tone, lesion-only tone, and lesion-skin contrast. The following subsections describe the tone computation and the distance measures.

\subsection{Distribution-based Skin Tone Measurement preserving unevenness and nuance}
We quantify skin tone using the Individual Typology Angle (ITA) (Equation \ref{eq:ita}), a standard colorimetric proxy for skin pigmentation computed in CIELab space (Equation \ref{eq:ita} adapted from \cite{wu2020utilization}). 

\begin{equation}
\label{eq:ita}
    ITA=\frac{ \arctan\left ( \frac{L*-50}{b*} \right )\times 180}{\pi }
\end{equation}

Where $L*$ represents lightness and $b*$ represents the blue-yellow axis in the CIELab color space. Many studies convert ITA
to a single scalar and then assign discrete skin types, such as the Fitzpatrick scale \cite{fitzpatrick1988validity}. While such categorization is convenient for classification fairness analyses, it can discard pixel-level variation that may matter for segmentation. We therefore adopt the distribution-based approach of \cite{paxton2024measuring}, comparing the ITA pixel distribution to a fixed reference distribution. Unlike \cite{paxton2024measuring}, we use a uniform reference anchored at $-30^\circ$ (the lower bound of the darkest group) to improve stability and avoid tying the reference to an arbitrary tone.

\subsection{Skin Tone Distance Measures and Segmentation Performance Evaluation}
We compare six patterns of ITA distributions (Figure \ref{fig:main}(c)). Patterns 1-3 measure the
distance between a reference distribution and the ITA distribution for (i) the full image, (ii)
skin-only pixels, and (iii) lesion-only pixels. We compute these distances using the
Wasserstein distance (WD), which compares cumulative distribution functions and
preserves subtle distributional differences in color and texture (Equation (\ref{eq:wd})). WD uses the sign function ($\sigma$), as defined in Equation (\ref{eq:sign}), to determine the color direction. WD has been shown to capture similarity relationships that are closer to human perception compared to L1 and L2 distances \cite{rubner2000earth,bonneel2015sliced}. Patterns 4-6 quantify within-image contrasts by taking WD between pairs of distributions (e.g., lesion-only vs skin-only), capturing lesion-skin pigment contrast relevant to boundary visibility. 

\begin{equation}
\label{eq:wd}
       W_{1}(x_{0}, x_{i}) = \sigma(x_{0}, x_{i}) \int_{}^{}|F_{0}(x)-F_{i}(x)|dx
\end{equation}

\begin{equation}
\label{eq:sign}
\sigma(x_{0}, x_{i}) =
\begin{cases}
-1, & \tilde{x}_{0} \ge \tilde{x}_{i},\\
\;\;1, & \tilde{x}_{0} < \tilde{x}_{i},
\end{cases}
\qquad
\tilde{x}_{j} = F_{j}^{-1}(0.5), \quad j \in \{0,i\}.
\end{equation}

For example, for distance 6 in Figure \ref{fig:main}, $x_{0}$ and $x_{i}$ denote the set of ITA of skin-only and lesion-only pixels, respectively. Their CDFs are compared using the WD to capture fine-grained color differences at the pixel level. We analyze differences among the aforementioned six skin-tone patterns and segmentation model performance. Segmentation performance is evaluated using nine commonly employed metrics, as listed in Table \ref{tab:metrics}.

\begin{table}[htbp]
\centering
\begin{tabular}{l|l|l}
\hline
Acronym & Full name  & Evaluation Criteria \\ \hline
IoU & Intersection over Union  & \multirow{7}{*}{\parbox{3cm}{Higher values indicate\\ better performance}} \\
DC  & Dice Coefficient &                \\
ST  & Sensitivity (True Positive Rate) &  \\
SP  & Specificity &                     \\
PA  & Pixel Accuracy &                  \\
AUC & Area under the Curve &            \\
CK  & Cohen's Kappa &                   \\ \hline
FPR & False Positive Rate &  \multirow{2}{*}{\parbox{3cm}{Lower values indicate\\ better performance}} \\
FNR & False Negative Rate &             \\ \hline
\end{tabular}
\caption{Segmentation Performance Evaluation Metrics}
\label{tab:metrics}
\end{table}

\section{Experimental Setup}
\subsection{Datasets}
We used two publicly available datasets: HAM for 7-class (Nevi (NV), Melanoma (MEL), Benign keratosis (BKL), Dermatofibroma (DF), Basal cell carcinoma (BCC), Vascular (VAS), and Actinic Keratoses (AK)) classification, and ISIC for binary classification (melanoma versus non-melanoma) to focus on the clinically significant distinction. HAM was divided into three parts: training, validation, and testing (60\%, 20\%, 20\%). For ISIC, we used its pre-split setup. All images are resized to 224 x 224 pixels, which is the optimal size for many pre-trained neural networks.

\subsection{Hair Removal}\label{sec:hair_removal}
Some dermoscopic images contain artifacts, such as hair, that can interfere with accurately capturing the true skin color \cite{corbin2023assessing}. We therefore applied a standard hair-removal pipeline prior to tone computation: (1) Noise reduction: An opening process was performed using a 3x3 pixel kernel to eliminate small noise. The kernel, a structural element used in morphological operations, is a matrix that focuses on the local area. (2) Hair and Shadow Emphasis: The Black Hat \cite{suiccmez2023detection} method with an 8x8 kernel was applied to highlight hair and shadow areas. (3) Contrast Limited Adaptive Histogram Equalization was applied to enhance contrast and facilitate hair detection. (4) Hair Mask Generation: Binary thresholding was employed to create masks, where hair appeared in white and the background in black. The kernel size and threshold values used in the morphological method were determined by visualizing the sample.

\subsection{Semantic Segmentation Model Selection}\label{sec:semseg_models}
UNET, DeepLabV3, and DINOv2 were selected to cover both established CNN architectures and a recent Vision Transformer backbone. UNET is widely used as a baseline for dermatological lesion segmentation \cite{mirikharaji2023survey,tanveer2024comprehensive}, and DeepLabV3 is a strong CNN-based alternative that often outperforms UNET in medical segmentation \cite{yao2024cnn}. DINOv2 has shown strong transfer to medical imaging via self-supervised pretraining on large-scale natural images \cite{huang2024comparative}. UNET and DeepLabV3 use ResNet50 backbones pretrained on ImageNet \cite{deng2009imagenet} and COCO/Pascal VOC subsets \cite{lin2014microsoft}\cite{everingham2010pascal}, respectively; DINOv2 used a ViT-B14 backbone pretrained on sets of natural image datasets. All models were fine-tuned for up to 50 epochs, selecting the checkpoint with the lowest validation Dice loss. Learning rates were 1e-5 (UNET/DeepLabV3) and 1e-6 (DINOv2), with linear warm-up over the first five epochs followed by scheduled decay.

\section{Results} \label{sec:results}
\begin{figure*}
    \centering
    \includegraphics[width=1\linewidth]{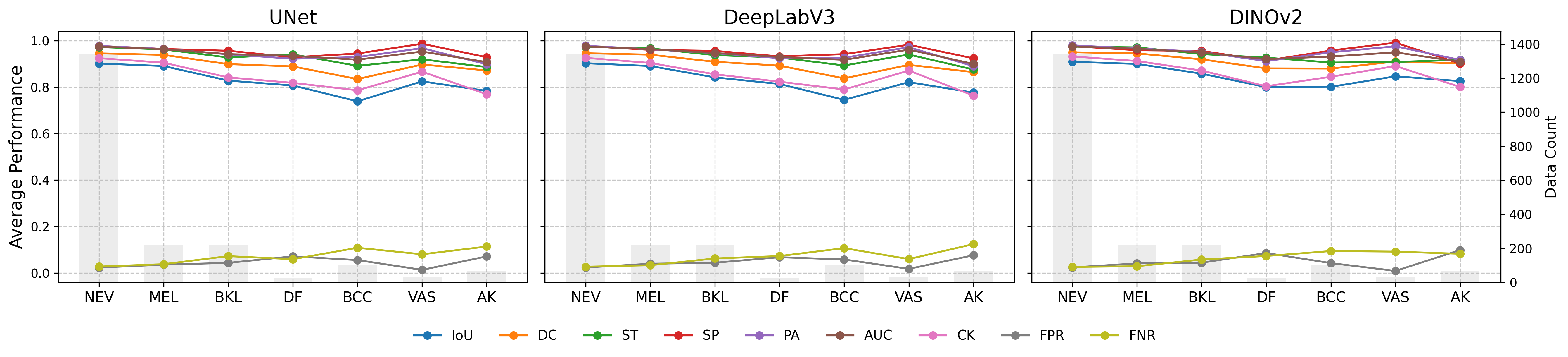}
    \caption{General Performance of each Class with HAM: The figure on the left shows metrics for evaluating prediction accuracy, and the higher the accuracy, the higher the value. In contrast, the figure on the right shows the error rate, and the lower the score, the better the model's performance.}
    \label{fig:class_performance_ham}
\end{figure*}

Table \ref{tab:performance} summarizes overall performance of each model. Across both datasets, IoU and CK are slightly lower than PA, sensitivity, and specificity, but all models achieved high accuracy (higher than 93\% in HAM and 83\% in ISIC) and low error rates, demonstrating excellent overall performance. Figure \ref{fig:class_performance_ham} reports disease-level performance on HAM and shows consistent trends across architectures, suggesting that performance variation is
driven largely by lesion characteristics rather than model family. Notably, classes such as BCC and AK show lower accuracy despite having more samples than DF and VAS, indicating that sample count alone does not explain class-wise difficulty. Similar trends are observed on ISIC.

\begin{table*}[htbp]
\centering
\small
\resizebox{0.8\textwidth}{!}{
\begin{tabular}{l|ll|lllllllll}
\hline
DataSet & Model & Epoch & IoU  & DC   & ST   & SP   & PA   & AUC  & CK   & FNR  & FPR  \\ \hline
\multirow{3}{*}{HAM}      
& UNET     & 24 & 0.88 & 0.93 & \textbf{0.96} & \textbf{0.97} & \textbf{0.97} & \textbf{0.96} & 0.90 & \textbf{0.04} & \textbf{0.03} \\ 
& DeepLab & 19 & 0.88 & 0.93 & \textbf{0.96} & \textbf{0.97} & \textbf{0.97} & \textbf{0.96} & 0.90 & \textbf{0.04} & \textbf{0.03} \\ 
& DinoV2  & 19 & 0.89 & 0.94 & \textbf{0.96} & \textbf{0.97} & \textbf{0.97} & \textbf{0.97} & 0.91 & \textbf{0.04} & \textbf{0.03} \\ \hline
\multirow{3}{*}{ISIC} 
& UNET     & 43 & 0.73 & 0.83 & 0.83 & \textbf{0.97} & 0.93 & 0.90 & 0.78 & 0.17 & \textbf{0.03} \\ 
& DeepLab & 49 & 0.73 & 0.83 & 0.83 & \textbf{0.97} & 0.93 & 0.90 & 0.77 & 0.17 & \textbf{0.03} \\ 
& DinoV2  & 42 & 0.75 & 0.84 & 0.84 & \textbf{0.97} & 0.94 & 0.91 & 0.79 & 0.16 & \textbf{0.03} \\ \hline
\end{tabular}}
\caption{General Performance of each Model: Bold values indicate metrics with high performance, defined as mean scores above 95\% for metrics where higher is better and below 5\% for metrics where lower is better.}
\label{tab:performance}
\end{table*}

\subsection{Patient's Skin Tone Distribution}
\begin{figure*}
    \centering
    \includegraphics[width=1\linewidth]{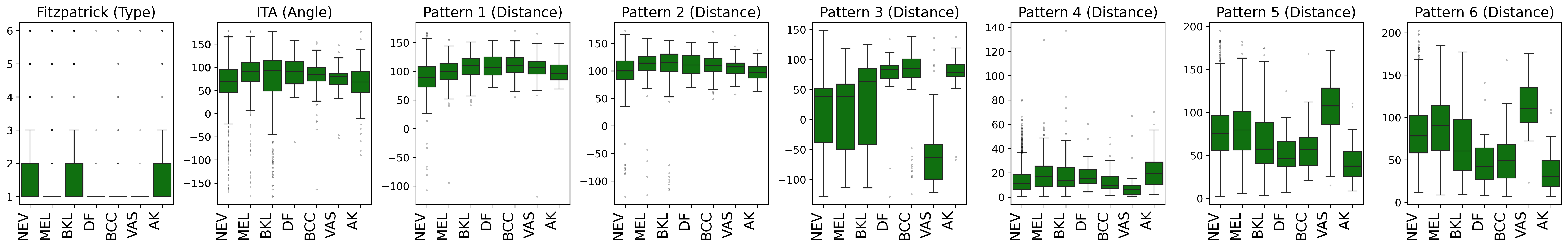}
    \caption{Skin Color Distribution Skin or Lesion Color Each Class: The eight figures in the upper row show the distribution of each class based on the evaluation of the colors of each image in the HAM. Each figure follows the measure methods for color patterns in Table \ref{fig:main} and shows trends in the distribution of color patterns in the dataset.}
    \label{fig:skin_color}
\end{figure*}

Figure \ref{fig:skin_color} summarizes skin tone distributions in HAM under several measurement
approaches. The first box plot shows that using Fitzpatrick categories, most cases fall into types 1-2, concentrating several classes (e.g., NV, BKL, AK), and limiting the extent to which tone effects can be
assessed via discrete grouping. Because most samples fall into Fitzpatrick types I–II, the statistical power to evaluate performance differences across darker skin tones is limited. Therefore, any conclusions regarding tone-dependent behaviour must be interpreted cautiously.

The second box plot shows the effect of ITA calculated as a single scalar. Measurements based on distance from the reference color (WD) show that the distribution for each condition is confined to a relatively narrow range across patterns 1-3. Patterns 4-6, which incorporate lesion-skin contrast, produce substantially wider distributions. This widening indicates that lesion pigmentation and lesion–skin contrast can dominate whole-image and skin-only tone proxies, motivating contrast-aware analysis. Comparable behavior is observed in ISIC.

\begin{figure*}
    \centering
    \includegraphics[width=1\linewidth]{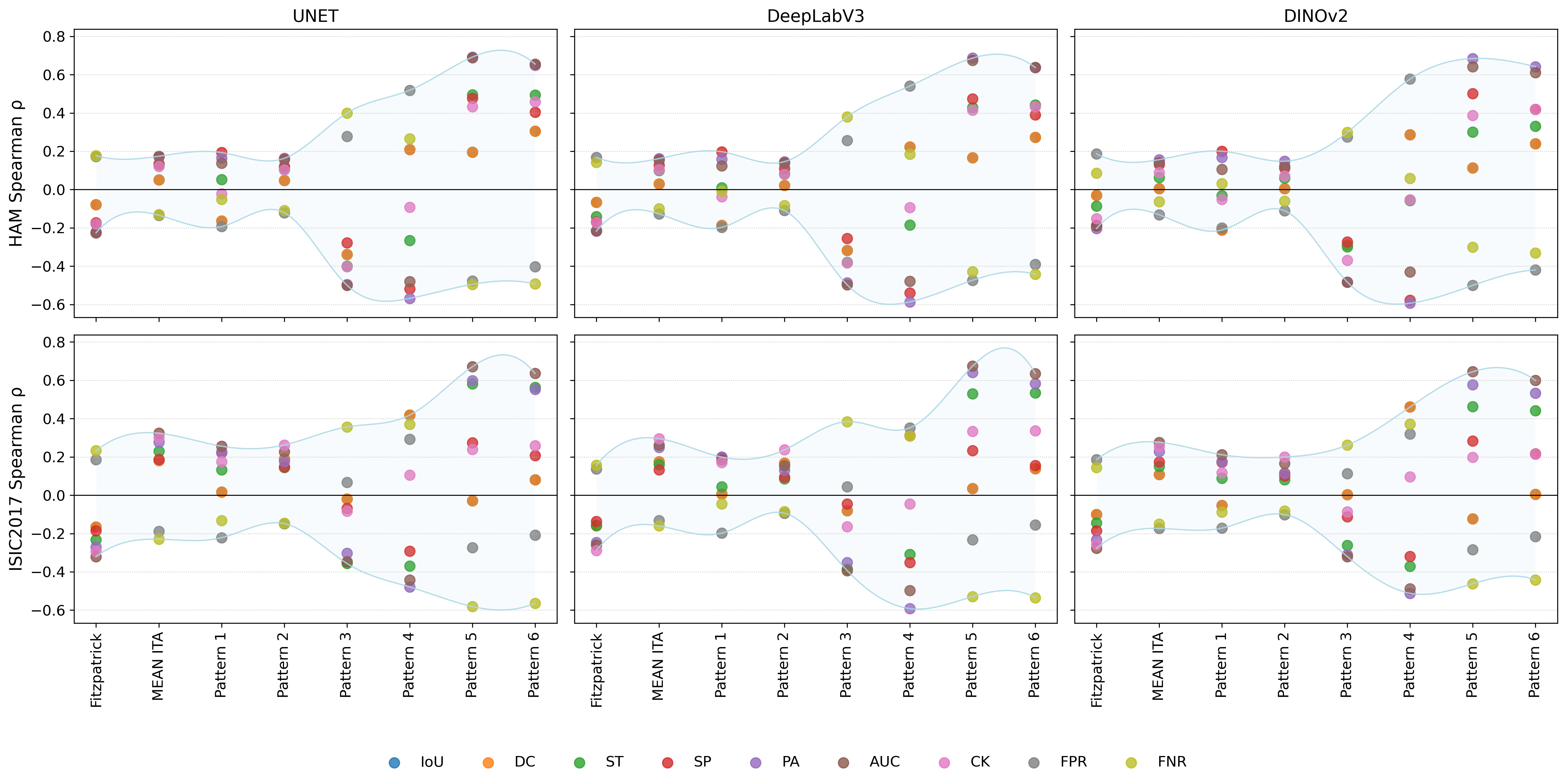}
    \caption{Correlation Ranges between Skin Color Measures and Segmentation Performance Across Datasets and Models: This figure presents Spearman rank correlations between each segmentation performance metric and skin-color distance measures across different comparison patterns. The light blue band visualizes the minimum–maximum range of correlations across metrics.}
    \label{fig:correlation}
\end{figure*}

\subsection{Influence of Skin Tone on Segmentation Performance}\label{sec:correlation}
Prior work reports reduced performance on darker skin tones in related dermatological
tasks \cite{kinyanjui2019estimating,alipour2024skin,rezk2022improving,benvcevic2024understanding}.Across models and datasets, global tone measures (Fitzpatrick grouping and mean ITA) show how weak association with segmentation performance (|$\rho$| < 0.2 in most metrics), and it is suggesting that tone alone explains little variance in segmentation accuracy within the available data range. This is shown in Figure \ref{fig:correlation} which presents Spearman's rank correlations. Likewise, WD measures focused on whole-image tone or skin-only tone (Patterns 1 and 2) show limited correlation with segmentation metrics across models. In contrast, when focusing on the area of skin lesions, a strong negative correlation was observed between the color of the lesion area and PA and AUC.

In HAM, across models, the skin-to-lesion color gap was correlated with segmentation performance across all metrics. A similar trend was observed for the overall-image–lesion gap, with correlations for most metrics except for IoU and DC. In ISIC, consistent with HAM, lesion pigment exhibited moderate negative correlations with performance across all models. For PA and AUC, performance increased as the overall-image–lesion color gap increased, yielding strong positive correlations. Additionally, ISIC showed a strong negative correlation between the overall-image-skin color gap and performance.

\begin{table}[]
\centering
\resizebox{1\columnwidth}{!}{
\begin{tabular}{ll|lll|lll|lll}
\hline
                           &        & \multicolumn{3}{l|}{UNET} & \multicolumn{3}{l|}{DeepLabV3} & \multicolumn{3}{l}{DINOv2} \\
& Metric & Corr   & L & H & Corr     & L   & H   & Corr   & L  & H  \\ \hline
\multirow{9}{*}{\rotatebox{90}{Pattern 4}} & IoU    & 0.21   & 0.16   & 0.25   & 0.22     & 0.18     & 0.27     & 0.29   & 0.24    & 0.33    \\
                           & DC     & 0.21   & 0.17   & 0.25   & 0.22     & 0.18     & 0.26     & 0.29   & 0.25    & 0.33    \\
                           & ST     & -0.27  & -0.31  & -0.22  & -0.19    & -0.23    & -0.14    & -0.06  & -0.10   & -0.01   \\
                           & SP     & -0.52  & -0.55  & -0.48  & -0.54    & -0.57    & -0.50    & -0.58  & -0.61   & -0.54   \\
                           & PA     & -0.57  & -0.60  & -0.53  & -0.59    & -0.62    & -0.56    & -0.59  & -0.62   & -0.56   \\
                           & AUC    & -0.48  & -0.52  & -0.44  & -0.48    & -0.52    & -0.44    & -0.43  & -0.47   & -0.39   \\
                           & CK     & -0.09  & -0.14  & -0.04  & -0.09    & -0.14    & -0.05    & -0.05  & -0.10   & -0.01   \\
                           & FPR    & 0.52   & 0.48   & 0.55   & 0.54     & 0.50     & 0.57     & 0.58   & 0.54    & 0.61    \\
                           & FNR    & 0.27   & 0.22   & 0.30   & 0.19     & 0.14     & 0.23     & 0.06   & 0.02    & 0.10    \\ \hline
\multirow{9}{*}{\rotatebox{90}{Pattern 5}} & IoU    & 0.20   & 0.15   & 0.24   & 0.17     & 0.12     & 0.21     & 0.11   & 0.07    & 0.16    \\
                           & DC     & 0.20   & 0.15   & 0.24   & 0.17     & 0.12     & 0.21     & 0.11   & 0.07    & 0.16    \\
                           & ST     & 0.50   & 0.46   & 0.53   & 0.43     & 0.39     & 0.47     & 0.30   & 0.26    & 0.34    \\
                           & SP     & 0.48   & 0.44   & 0.51   & 0.47     & 0.44     & 0.51     & 0.50   & 0.46    & 0.54    \\
                           & PA     & 0.69   & 0.66   & 0.72   & 0.69     & 0.66     & 0.71     & 0.68   & 0.66    & 0.71    \\
                           & AUC    & 0.69   & 0.66   & 0.71   & 0.67     & 0.65     & 0.70     & 0.64   & 0.61    & 0.67    \\
                           & CK     & 0.43   & 0.40   & 0.47   & 0.41     & 0.38     & 0.45     & 0.39   & 0.35    & 0.43    \\
                           & FPR    & -0.48  & -0.52  & -0.44  & -0.47    & -0.51    & -0.44    & -0.50  & -0.54   & -0.46   \\
                           & FNR    & -0.50  & -0.53  & -0.46  & -0.43    & -0.46    & -0.39    & -0.30  & -0.34   & -0.26   \\ \hline
\multirow{9}{*}{\rotatebox{90}{Pattern 6}} & IoU    & 0.30   & 0.26   & 0.35   & 0.27     & 0.23     & 0.32     & 0.24   & 0.19    & 0.28    \\
                           & DC     & 0.30   & 0.27   & 0.35   & 0.27     & 0.23     & 0.31     & 0.24   & 0.20    & 0.28    \\
                           & ST     & 0.49   & 0.46   & 0.53   & 0.44     & 0.41     & 0.48     & 0.33   & 0.29    & 0.37    \\
                           & SP     & 0.40   & 0.36   & 0.44   & 0.39     & 0.35     & 0.43     & 0.42   & 0.38    & 0.46    \\
                           & PA     & 0.65   & 0.62   & 0.67   & 0.64     & 0.61     & 0.66     & 0.64   & 0.61    & 0.67    \\
                           & AUC    & 0.66   & 0.63   & 0.68   & 0.64     & 0.61     & 0.66     & 0.61   & 0.58    & 0.64    \\
                           & CK     & 0.46   & 0.42   & 0.49   & 0.43     & 0.40     & 0.47     & 0.42   & 0.38    & 0.46    \\
                           & FPR    & -0.40  & -0.44  & -0.36  & -0.39    & -0.43    & -0.35    & -0.42  & -0.46   & -0.38   \\
                           & FNR    & -0.49  & -0.53  & -0.46  & -0.44    & -0.48    & -0.40    & -0.33  & -0.37   & -0.29  \\ \hline
\end{tabular}
}
\caption{Confidence intervals of Spearman correlation coefficients for Patterns 4-6}
\label{tab:ci_ham}
\end{table}

To assess robustness of the observed correlations, we computed 95\% confidence intervals for Spearman’s rank correlations using 1000 bootstrap resamples. Table \ref{tab:ci_ham} reports results for Patterns 4–6 in HAM. L denotes the lower bound of the CI, and H denotes the 97.5\% (upper bound). Correlation signs are preserved across the confidence intervals, and interval widths remain small (approximately 0.07–0.075 across architectures), supporting the stability of the estimated correlations. All reported p-values are less than 0.001, indicating statistically reliable associations. Comparable trends were also observed in ISIC.

\subsection{Influence of Skin Tone on Each Disease}\label{sec:class_correlation}
\begin{figure*}
    \centering
    \includegraphics[width=1\linewidth]{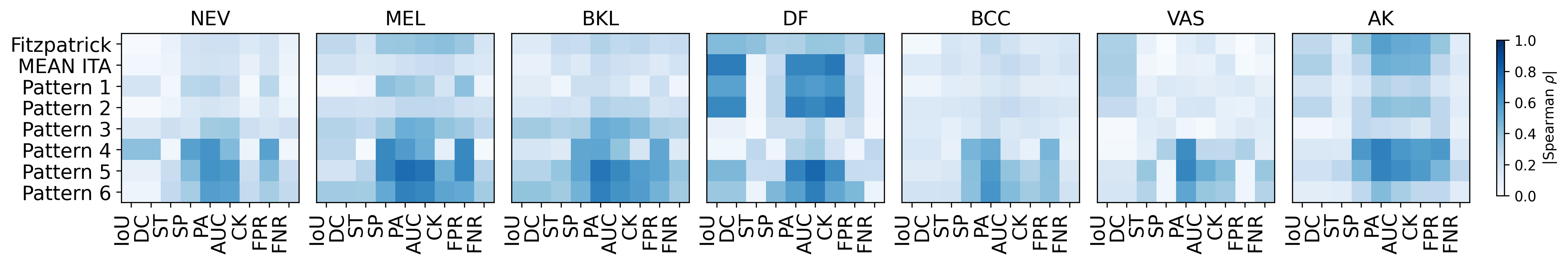}
    \caption{DINOv2 Absolute Spearman correlations between skin tone and segmentation performance metrics across seven disease classes}
    \label{fig:class_correlation_heatmap}
\end{figure*}

Disease-stratified analysis in HAM further supports the primacy of contrast over discrete tone grouping. When stratified by Fitzpatrick categories, correlations between tone and segmentation metrics are inconsistent across diseases and metrics (Figure \ref{fig:class_correlation_heatmap}), suggesting that discrete tone binning does not reliably explain performance variation at the disease level. In contrast, contrast-sensitive WD Patterns 4–6 showed large magnitudes of Spearman’s rank absolute correlation across multiple diseases, including cancerous (e.g., MEL, BCC) and low-performance diseases (e.g., DF, BCC, AK), as well as across multiple evaluation metrics. These patterns are observed consistently across UNET, DeepLabV3, and DINOv2,  suggesting that segmentation performance depends on the contrast of lesions.

We further evaluated malignant classes (MEL, BCC) using bootstrap confidence intervals for moderate-to-strong correlations (±0.4) (see Table \ref{tab:ci_local}). The reported intervals are relatively narrow and all corresponding p-values are less than 0.001, supporting the reliability of contrast–performance associations in these clinically important subsets.

\begin{figure}
    \centering
    \includegraphics[width=1\linewidth]{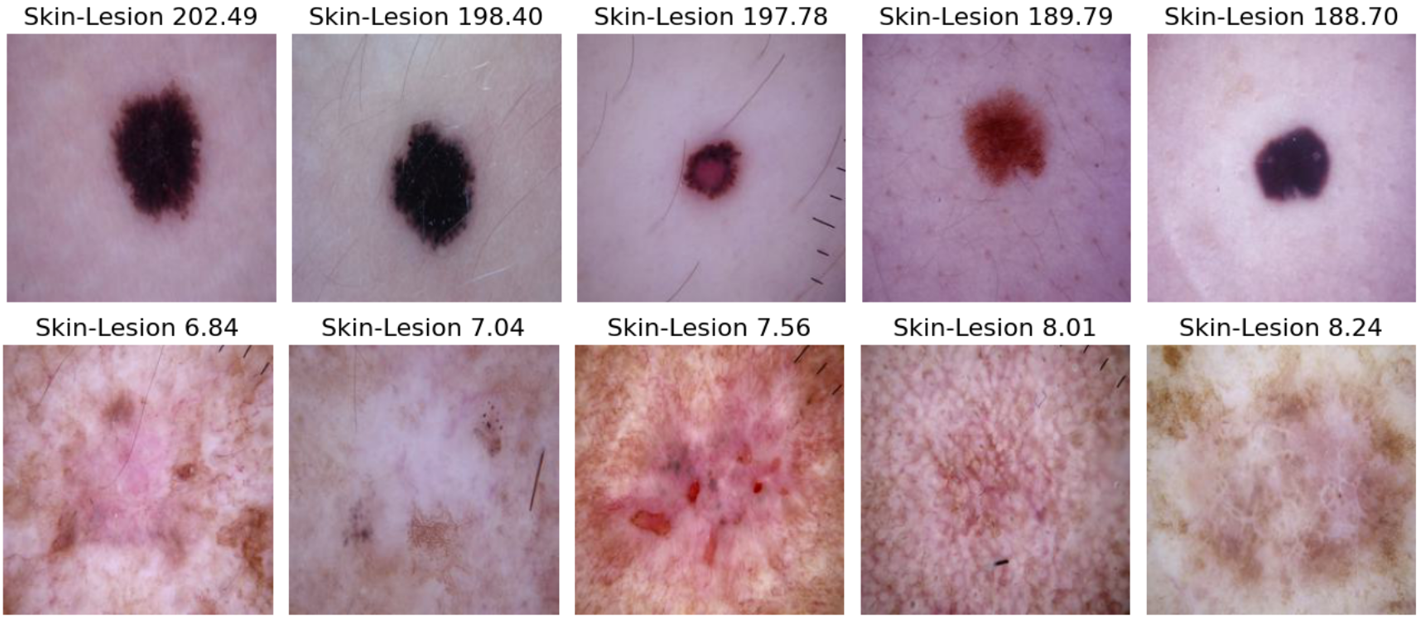}
    \caption{Sample images measured for skin tone nuances using Pattern 6: The difference between skin pixels and lesion areas can be clearly captured and quantified.}
    \label{fig:samples}
\end{figure}

Finally, Figure \ref{fig:samples} provides qualitative examples consistent with the quantitative results: high-contrast cases exhibit visually distinct boundaries, whereas low-contrast cases show ambiguous boundaries that align with increased segmentation difficulty.

\begin{table}[]
\centering
\resizebox{1\columnwidth}{!}{
\begin{tabular}{lll|lll|lll|lll}
\hline
& & & \multicolumn{3}{|l}{UNET} & \multicolumn{3}{|l}{DeepLabV3} & \multicolumn{3}{|l}{DINOv2} \\
& & Metric & Corr   & L      & H      & Corr     & L        & H       & Corr    & L       & H      \\ \hline
\multirow{18}{*}{\rotatebox{90}{MEL}} & \multirow{4}{*}{\rotatebox{90}{Pattern 4}} & SP     & -0.59  & -0.67  & -0.50  & -0.64    & -0.71    & -0.55   & -0.66   & -0.73   & -0.57  \\
                    &                            & PA     & -0.52  & -0.61  & -0.40  & -0.58    & -0.67    & -0.48   & -0.58   & -0.66   & -0.48  \\
                    &                            & AUC    & -0.48  & -0.59  & -0.37  & -0.51    & -0.61    & -0.40   & -0.49   & -0.58   & -0.39  \\
                    &                            & FPR    & 0.59   & 0.50   & 0.68   & 0.64     & 0.56     & 0.71    & 0.66    & 0.59    & 0.73   \\
                    & \multirow{7}{*}{\rotatebox{90}{Pattern 5}} & ST     & 0.57   & 0.47   & 0.66   & 0.42     & 0.29     & 0.53    & 0.30    & 0.16    & 0.42   \\
                    &                            & SP     & 0.56   & 0.45   & 0.66   & 0.65     & 0.56     & 0.72    & 0.67    & 0.58    & 0.74   \\
                    &                            & PA     & 0.74   & 0.67   & 0.80   & 0.77     & 0.70     & 0.82    & 0.77    & 0.71    & 0.82   \\
                    &                            & AUC    & 0.78   & 0.72   & 0.83   & 0.79     & 0.73     & 0.84    & 0.74    & 0.66    & 0.79   \\
                    &                            & CK     & 0.49   & 0.38   & 0.58   & 0.51     & 0.40     & 0.61    & 0.49    & 0.38    & 0.59   \\
                    &                            & FPR    & -0.56  & -0.66  & -0.46  & -0.65    & -0.72    & -0.56   & -0.67   & -0.75   & -0.58  \\
                    &                            & FNR    & -0.57  & -0.66  & -0.47  & -0.42    & -0.53    & -0.30   & -0.30   & -0.41   & -0.16  \\
                    & \multirow{7}{*}{\rotatebox{90}{Pattern 6}} & ST     & 0.59   & 0.50   & 0.67   & 0.44     & 0.31     & 0.55    & 0.36    & 0.23    & 0.47   \\
                    &                            & SP     & 0.43   & 0.30   & 0.54   & 0.52     & 0.41     & 0.62    & 0.52    & 0.42    & 0.62   \\
                    &                            & PA     & 0.67   & 0.59   & 0.75   & 0.67     & 0.56     & 0.74    & 0.68    & 0.60    & 0.75   \\
                    &                            & AUC    & 0.71   & 0.63   & 0.78   & 0.70     & 0.62     & 0.77    & 0.67    & 0.58    & 0.73   \\
                    &                            & CK     & 0.54   & 0.43   & 0.64   & 0.53     & 0.42     & 0.63    & 0.54    & 0.44    & 0.63   \\
                    &                            & FPR    & -0.43  & -0.55  & -0.32  & -0.52    & -0.61    & -0.42   & -0.52   & -0.61   & -0.42  \\
                    &                            & FNR    & -0.59  & -0.67  & -0.50  & -0.44    & -0.54    & -0.31   & -0.36   & -0.47   & -0.23  \\ \hline
\multirow{15}{*}{\rotatebox{90}{BCC}} & \multirow{3}{*}{\rotatebox{90}{Pattern 4}} & SP     & -0.17  & -0.34  & 0.02   & -0.23    & -0.43    & -0.04   & -0.47   & -0.63   & -0.30  \\
                    &                            & PA     & -0.43  & -0.58  & -0.25  & -0.44    & -0.61    & -0.24   & -0.51   & -0.66   & -0.32  \\
                    &                            & FPR    & 0.17   & -0.02  & 0.34   & 0.23     & 0.03     & 0.41    & 0.47    & 0.29    & 0.62   \\
                    & \multirow{6}{*}{\rotatebox{90}{Pattern 5}} & ST     & 0.46   & 0.30   & 0.60   & 0.39     & 0.21     & 0.54    & 0.15    & -0.03   & 0.34   \\
                    &                            & SP     & 0.09   & -0.10  & 0.27   & 0.18     & -0.02    & 0.36    & 0.42    & 0.24    & 0.57   \\
                    &                            & PA     & 0.45   & 0.26   & 0.60   & 0.47     & 0.28     & 0.63    & 0.59    & 0.43    & 0.70   \\
                    &                            & AUC    & 0.44   & 0.27   & 0.59   & 0.41     & 0.23     & 0.56    & 0.40    & 0.21    & 0.56   \\
                    &                            & FPR    & -0.09  & -0.27  & 0.09   & -0.18    & -0.37    & 0.02    & -0.42   & -0.57   & -0.25  \\
                    &                            & FNR    & -0.46  & -0.59  & -0.30  & -0.39    & -0.54    & -0.21   & -0.15   & -0.32   & 0.02   \\
                    & \multirow{6}{*}{\rotatebox{90}{Pattern 6}} & ST     & 0.46   & 0.30   & 0.60   & 0.43     & 0.26     & 0.57    & 0.19    & 0.01    & 0.35   \\
                    &                            & SP     & 0.11   & -0.08  & 0.32   & 0.14     & -0.07    & 0.36    & 0.42    & 0.24    & 0.56   \\
                    &                            & PA     & 0.47   & 0.31   & 0.62   & 0.48     & 0.30     & 0.64    & 0.61    & 0.45    & 0.73   \\
                    &                            & AUC    & 0.46   & 0.28   & 0.62   & 0.44     & 0.26     & 0.59    & 0.44    & 0.26    & 0.59   \\
                    &                            & FPR    & -0.11  & -0.31  & 0.09   & -0.14    & -0.33    & 0.06    & -0.42   & -0.57   & -0.26  \\
                    &                            & FNR    & -0.46  & -0.60  & -0.29  & -0.43    & -0.58    & -0.26   & -0.19   & -0.36   & -0.01 \\ \hline
\end{tabular}
}
\caption{Confidence intervals of moderate-to-strong Spearman correlation coefficients for malignant skin lesions (Pattern 4-6)}
\label{tab:ci_local}
\end{table}

\section{Discussion and Future Work}
\subsection{Discussion}
This section addresses the research questions using the empirical findings. 

For  \textbf{RQ1}, global skin tone measures (Fitzpatrick grouping and mean ITA) show a weak association with lesion segmentation performance. This result holds across three segmentation families (UNET, DeepLabV3, DINOv2) and across multiple evaluation metrics, within the range of tones represented in HAM10000 and ISIC2017. 

For \textbf{RQ2} lesion–skin pigment contrast, quantified using WD between lesion-only and skin-only ITA distributions, shows consistent and statistically supported associations with segmentation quality. In particular, lower lesion–skin contrast is associated with reduced accuracy and increased error rates (Table \ref{tab:ci_ham}), and this effect persists across architectures. These findings provide quantitative support for prior qualitative observations that low contrast and boundary ambiguity are dominant failure modes in dermoscopic lesion segmentation \cite{mirikharaji2023survey,khanra2022survey}. While previous work has noted contrast as a plausible contributor (often via visual inspection), few studies have operationalized contrast as a continuous, within-image variable suitable for systematic auditing \cite{benvcevic2024understanding}. The distribution-based contrast measures used here address this gap by preserving pixel-level gradation and enabling contrast to be compared across images without collapsing tone into coarse categories.

For \textbf{RQ3}, contrast measures also align with error behavior: low-contrast conditions are associated with higher FNR and FPR (Table III), consistent with boundary ambiguity leading to both under- and over-segmentation. The malignant-class analysis (Table \ref{tab:ci_local}) indicates that these relationships are not confined to benign categories and can be observed in clinically important subsets (MEL, BCC), reinforcing the relevance of contrast-aware evaluation for safety-critical use.

Collectively, these results suggest that segmentation fairness and reliability efforts should prioritize low-contrast lesion handling rather than relying primarily on discrete tone binning. Practical implications include: 

\begin{enumerate}
    \item contrast-stratified evaluation reporting,
    \item targeted sampling or augmentation to increase representation of low-contrast cases, and
    \item the use of continuous contrast scores as an auditing or monitoring signal.
\end{enumerate}

In clinical workflows, contrast-based flags could be used to identify cases more likely to yield uncertain boundaries, supporting safer review pathways when segmentation outputs feed downstream classification or triage.

\subsection{Future Work}
We plan to utilise these findings to explore adaptive control of feature fusion at the decoder stage for low-contrast images detected by WD. We also investigate removing structures incapable of incorporating contrast through pruning techniques.

\subsection{Limitations}
Skin colour can be influenced by lighting conditions and camera devices \cite{zhou2024auto,mbatha2024skin}; however, our method mitigated this effect by using relative measurements rather than directly applying ITA, as in previous research. The assessment of fairness in this study is limited to the range of skin tones included within the HAM10000 and ISIC datasets. These datasets exhibit bias in skin tone distribution, with darker skin tones potentially underrepresented. Therefore, generalizing the results of this study to a broader range of skin tones may require additional validation.

\section{Conclusion}
This study examined how global skin tone measures and lesion-skin pigment contrast relate to dermoscopic lesion segmentation using UNET, DeepLabV3, and DINOv2 across the HAM and ISIC datasets. Across architectures and metrics, segmentation performance was more strongly correlated with lesion-skin contrast than global skin tone alone. We further showed that discrete tone categorization (e.g., Fitzpatrick) can obscure within-image gradations that are salient for pixelwise analysis, whereas distribution-based representations enable quantitative characterization of contrast–performance relationships.  

From a patient-safety perspective, low-contrast cases are more likely to yield boundary ambiguity and under-segmentation, which can propagate downstream to classification and triage errors. Practically, these results motivate contrast-aware evaluation and mitigation strategies (e.g., targeted sampling, augmentation, or uncertainty-triggered review) as a pathway to more reliable and equitable dermatological pipelines. Because darker tones are under-represented in these public datasets, validating these findings on more diverse cohorts remains an important next step.

\section*{Acknowledgment}
The authors would like to thank the Data Science, Artificial Intelligence, and Modelling Institute at the University of Hull for their support. The authors extend their gratitude to Dr. Jun-ya Norimatsu at ALINEAR Corp. for technical advice with the experiments.

\section*{Reference}
\bibliographystyle{IEEEtran}

\begin{thebibliography}{10}
\providecommand{\url}[1]{#1}
\csname url@samestyle\endcsname
\providecommand{\newblock}{\relax}
\providecommand{\bibinfo}[2]{#2}
\providecommand{\BIBentrySTDinterwordspacing}{\spaceskip=0pt\relax}
\providecommand{\BIBentryALTinterwordstretchfactor}{4}
\providecommand{\BIBentryALTinterwordspacing}{\spaceskip=\fontdimen2\font plus
\BIBentryALTinterwordstretchfactor\fontdimen3\font minus \fontdimen4\font\relax}
\providecommand{\BIBforeignlanguage}[2]{{%
\expandafter\ifx\csname l@#1\endcsname\relax
\typeout{** WARNING: IEEEtran.bst: No hyphenation pattern has been}%
\typeout{** loaded for the language `#1'. Using the pattern for}%
\typeout{** the default language instead.}%
\else
\language=\csname l@#1\endcsname
\fi
#2}}
\providecommand{\BIBdecl}{\relax}
\BIBdecl

\bibitem{kasmi2023sharprazor}
R.~Kasmi, J.~Hagerty, R.~Young, N.~Lama, J.~Nepal, J.~Miinch, W.~Stoecker, and R.~J. Stanley, ``Sharprazor: automatic removal of hair and ruler marks from dermoscopy images,'' \emph{Skin Research and Technology}, vol.~29, no.~4, p. e13203, 2023.

\bibitem{rizzi2020skin}
M.~Rizzi and C.~Guaragnella, ``Skin lesion segmentation using image bit-plane multilayer approach,'' \emph{Applied Sciences}, vol.~10, no.~9, p. 3045, 2020.

\bibitem{lopez2022interpretable}
J.~Lopez-Labraca, I.~Gonzalez-Diaz, F.~Diaz-de Maria, and A.~Fueyo-Casado, ``An interpretable cnn-based cad system for skin lesion diagnosis,'' \emph{Artificial Intelligence in Medicine}, vol. 132, p. 102370, 2022.

\bibitem{pakzad2022circle}
A.~Pakzad, K.~Abhishek, and G.~Hamarneh, ``Circle: Color invariant representation learning for unbiased classification of skin lesions,'' in \emph{European Conference on Computer Vision}.\hskip 1em plus 0.5em minus 0.4em\relax Springer, 2022, pp. 203--219.

\bibitem{kinyanjui2019estimating}
N.~M. Kinyanjui, T.~Odonga, C.~Cintas, N.~C. Codella, R.~Panda, P.~Sattigeri, and K.~R. Varshney, ``Estimating skin tone and effects on classification performance in dermatology datasets,'' \emph{arXiv preprint arXiv:1910.13268}, 2019.

\bibitem{rezk2022improving}
E.~Rezk, M.~Eltorki, W.~El-Dakhakhni \emph{et~al.}, ``Improving skin color diversity in cancer detection: deep learning approach,'' \emph{JMIR dermatology}, vol.~5, no.~3, p. e39143, 2022.

\bibitem{montoya2024towards}
L.~N. Montoya, J.~S. Roberts, and B.~S. Hidalgo, ``Towards fairness in ai for melanoma detection: Systemic review and recommendations,'' \emph{arXiv preprint arXiv:2411.12846}, 2024.

\bibitem{ronneberger2015u}
O.~Ronneberger, P.~Fischer, and T.~Brox, ``U-net: Convolutional networks for biomedical image segmentation,'' in \emph{Medical image computing and computer-assisted intervention--MICCAI 2015: 18th international conference, Munich, Germany, October 5-9, 2015, proceedings, part III 18}.\hskip 1em plus 0.5em minus 0.4em\relax Springer, 2015, pp. 234--241.

\bibitem{chen2017rethinking}
L.-C. Chen, ``Rethinking atrous convolution for semantic image segmentation,'' \emph{arXiv preprint arXiv:1706.05587}, 2017.

\bibitem{oquab2023dinov2}
M.~Oquab, T.~Darcet, T.~Moutakanni, H.~Vo, M.~Szafraniec, V.~Khalidov, P.~Fernandez, D.~Haziza, F.~Massa, A.~El-Nouby \emph{et~al.}, ``Dinov2: Learning robust visual features without supervision,'' \emph{arXiv preprint arXiv:2304.07193}, 2023.

\bibitem{tschandl2018ham10000}
P.~Tschandl, C.~Rosendahl, and H.~Kittler, ``The ham10000 dataset, a large collection of multi-source dermatoscopic images of common pigmented skin lesions. scientific data. 2018; 5: 180161,'' \emph{Search in}, vol.~2, 2018.

\bibitem{tschandl2020human}
P.~Tschandl, C.~Rinner, Z.~Apalla, G.~Argenziano, N.~Codella, A.~Halpern, M.~Janda, A.~Lallas, C.~Longo, J.~Malvehy \emph{et~al.}, ``Human--computer collaboration for skin cancer recognition,'' \emph{Nature Medicine}, vol.~26, no.~8, pp. 1229--1234, 2020.

\bibitem{codella2018skin}
N.~C. Codella, D.~Gutman, M.~E. Celebi, B.~Helba, M.~A. Marchetti, S.~W. Dusza, A.~Kalloo, K.~Liopyris, N.~Mishra, H.~Kittler \emph{et~al.}, ``Skin lesion analysis toward melanoma detection: A challenge at the 2017 international symposium on biomedical imaging (isbi), hosted by the international skin imaging collaboration (isic),'' in \emph{2018 IEEE 15th international symposium on biomedical imaging (ISBI 2018)}.\hskip 1em plus 0.5em minus 0.4em\relax IEEE, 2018, pp. 168--172.

\bibitem{gamage2024melanoma}
L.~Gamage, U.~Isuranga, D.~Meedeniya, S.~De~Silva, and P.~Yogarajah, ``Melanoma skin cancer identification with explainability utilizing mask guided technique,'' \emph{Electronics}, vol.~13, no.~4, p. 680, 2024.

\bibitem{saha2024yotransvit}
D.~K. Saha, A.~M. Joy, and A.~Majumder, ``Yotransvit: A transformer and cnn method for predicting and classifying skin diseases using segmentation techniques,'' \emph{Informatics in Medicine Unlocked}, vol.~47, p. 101495, 2024.

\bibitem{singh2022skinet}
R.~K. Singh, R.~Gorantla, S.~G.~R. Allada, and P.~Narra, ``Skinet: A deep learning framework for skin lesion diagnosis with uncertainty estimation and explainability,'' \emph{Plos one}, vol.~17, no.~10, p. e0276836, 2022.

\bibitem{mahbod2020effects}
A.~Mahbod, P.~Tschandl, G.~Langs, R.~Ecker, and I.~Ellinger, ``The effects of skin lesion segmentation on the performance of dermatoscopic image classification,'' \emph{Computer Methods and Programs in Biomedicine}, vol. 197, p. 105725, 2020.

\bibitem{minhas2020accurate}
K.~Minhas, T.~M. Khan, M.~Arsalan, S.~S. Naqvi, M.~Ahmed, H.~A. Khan, M.~A. Haider, and A.~Haseeb, ``Accurate pixel-wise skin segmentation using shallow fully convolutional neural network,'' \emph{IEEE Access}, vol.~8, pp. 156\,314--156\,327, 2020.

\bibitem{adegun2021deep}
A.~Adegun and S.~Viriri, ``Deep learning techniques for skin lesion analysis and melanoma cancer detection: a survey of state-of-the-art,'' \emph{Artificial Intelligence Review}, vol.~54, no.~2, pp. 811--841, 2021.

\bibitem{khan2023identifying}
S.~Khan, H.~Ali, and Z.~Shah, ``Identifying the role of vision transformer for skin cancer—a scoping review,'' \emph{Frontiers in Artificial Intelligence}, vol.~6, p. 1202990, 2023.

\bibitem{zhang2023recent}
J.~Zhang, F.~Zhong, K.~He, M.~Ji, S.~Li, and C.~Li, ``Recent advancements and perspectives in the diagnosis of skin diseases using machine learning and deep learning: A review,'' \emph{Diagnostics}, vol.~13, no.~23, p. 3506, 2023.

\bibitem{guo2022bias}
L.~N. Guo, M.~S. Lee, B.~Kassamali, C.~Mita, and V.~E. Nambudiri, ``Bias in, bias out: underreporting and underrepresentation of diverse skin types in machine learning research for skin cancer detection—a scoping review,'' \emph{Journal of the American Academy of Dermatology}, vol.~87, no.~1, pp. 157--159, 2022.

\bibitem{wen2022characteristics}
D.~Wen, S.~M. Khan, A.~J. Xu, H.~Ibrahim, L.~Smith, J.~Caballero, L.~Zepeda, C.~de~Blas~Perez, A.~K. Denniston, X.~Liu \emph{et~al.}, ``Characteristics of publicly available skin cancer image datasets: a systematic review,'' \emph{The Lancet Digital Health}, vol.~4, no.~1, pp. e64--e74, 2022.

\bibitem{pope2024skin}
J.~Pope, M.~Hassanuzzaman, M.~Sherpa, O.~Emara, A.~Joshi, and N.~Adhikari, ``Skin cancer machine learning model tone bias,'' \emph{arXiv preprint arXiv:2410.06385}, 2024.

\bibitem{abhari2023mitigating}
J.~Abhari and A.~Ashok, ``Mitigating racial biases for machine learning based skin cancer detection,'' in \emph{Proceedings of the Twenty-Fourth International Symposium on Theory, Algorithmic Foundations, and Protocol Design for Mobile Networks and Mobile Computing}, 2023, pp. 556--561.

\bibitem{xu2022achieving}
G.~Xu, Y.~Wu, J.~Hu, and Y.~Shi, ``Achieving fairness in dermatological disease diagnosis through automatic weight adjusting federated learning and personalization,'' \emph{arXiv preprint arXiv:2208.11187}, 2022.

\bibitem{fan2021fairness}
D.~Fan, Y.~Wu, and X.~Li, ``On the fairness of swarm learning in skin lesion classification,'' in \emph{Clinical Image-Based Procedures, Distributed and Collaborative Learning, Artificial Intelligence for Combating COVID-19 and Secure and Privacy-Preserving Machine Learning: 10th Workshop, CLIP 2021, Second Workshop, DCL 2021, First Workshop, LL-COVID19 2021, and First Workshop and Tutorial, PPML 2021, Held in Conjunction with MICCAI 2021, Strasbourg, France, September 27 and October 1, 2021, Proceedings 2}.\hskip 1em plus 0.5em minus 0.4em\relax Springer, 2021, pp. 120--129.

\bibitem{kong2024achieving}
Q.~Kong, C.-H. Chiu, D.~Zeng, Y.-J. Chen, T.-Y. Ho, J.~Hu, and Y.~Shi, ``Achieving fairness through channel pruning for dermatological disease diagnosis,'' in \emph{International Conference on Medical Image Computing and Computer-Assisted Intervention}.\hskip 1em plus 0.5em minus 0.4em\relax Springer, 2024, pp. 24--34.

\bibitem{wu2022fairprune}
Y.~Wu, D.~Zeng, X.~Xu, Y.~Shi, and J.~Hu, ``Fairprune: Achieving fairness through pruning for dermatological disease diagnosis,'' in \emph{International Conference on Medical Image Computing and Computer-Assisted Intervention}.\hskip 1em plus 0.5em minus 0.4em\relax Springer, 2022, pp. 743--753.

\bibitem{paxton2026enhancing}
K.~Paxton, K.~Aslansefat, D.~Thakker, Y.~Papadopoulos, and T.~Maslekar, ``Enhancing fairness in skin lesion classification for medical diagnosis using prune learning,'' \emph{IEEE Journal of Biomedical and Health Informatics}, 2026.

\bibitem{aayushman2024fair}
Aayushman, H.~Gaddey, V.~Mittal, M.~Chawla, and G.~R. Gupta, ``Fair and accurate skin disease image classification by alignment with clinical labels,'' in \emph{International Conference on Medical Image Computing and Computer-Assisted Intervention}.\hskip 1em plus 0.5em minus 0.4em\relax Springer, 2024, pp. 394--404.

\bibitem{du2022fairdisco}
S.~Du, B.~Hers, N.~Bayasi, G.~Hamarneh, and R.~Garbi, ``Fairdisco: Fairer ai in dermatology via disentanglement contrastive learning,'' in \emph{European Conference on Computer Vision}.\hskip 1em plus 0.5em minus 0.4em\relax Springer, 2022, pp. 185--202.

\bibitem{chiu2024achieve}
C.-H. Chiu, Y.-J. Chen, Y.~Wu, Y.~Shi, and T.-Y. Ho, ``Achieve fairness without demographics for dermatological disease diagnosis,'' \emph{Medical Image Analysis}, vol.~95, p. 103188, 2024.

\bibitem{bissoto2020debiasing}
A.~Bissoto, E.~Valle, and S.~Avila, ``Debiasing skin lesion datasets and models? not so fast,'' in \emph{Proceedings of the IEEE/CVF Conference on Computer Vision and Pattern Recognition Workshops}, 2020, pp. 740--741.

\bibitem{benvcevic2024understanding}
M.~Ben{\v{c}}evi{\'c}, M.~Habijan, I.~Gali{\'c}, D.~Babin, and A.~Pi{\v{z}}urica, ``Understanding skin color bias in deep learning-based skin lesion segmentation,'' \emph{Computer methods and programs in biomedicine}, vol. 245, p. 108044, 2024.

\bibitem{medelink}
\BIBentryALTinterwordspacing
M.~S.~R. Solution, ``Skin colorimeter cl440: Skin color measuring device,'' Mar 2025. [Online]. Available: \url{https://medelink.ca/research-devices/probes/sub-page-cl400/?srsltid=AfmBOop0J6waIAbionDUixfHmRpTnjZnC2RKpTR815a6GFrY1vdTJEXR}
\BIBentrySTDinterwordspacing

\bibitem{wu2020utilization}
Y.~Wu, T.~Tanaka, and M.~Akimoto, ``Utilization of individual typology angle (ita) and hue angle in the measurement of skin color on images,'' \emph{bioimages}, vol.~28, pp. 1--8, 2020.

\bibitem{fitzpatrick1988validity}
T.~B. Fitzpatrick, ``The validity and practicality of sun-reactive skin types i through vi,'' \emph{Archives of dermatology}, vol. 124, no.~6, pp. 869--871, 1988.

\bibitem{paxton2024measuring}
K.~Paxton, K.~Aslansefat, D.~Thakker, and Y.~Papadopoulos, ``Measuring ai fairness in a continuum maintaining nuances: A robustness case study,'' \emph{IEEE Internet Computing}, 2024.

\bibitem{rubner2000earth}
Y.~Rubner, C.~Tomasi, and L.~J. Guibas, ``The earth mover's distance as a metric for image retrieval,'' \emph{International journal of computer vision}, vol.~40, no.~2, pp. 99--121, 2000.

\bibitem{bonneel2015sliced}
N.~Bonneel, J.~Rabin, G.~Peyr{\'e}, and H.~Pfister, ``Sliced and radon wasserstein barycenters of measures,'' \emph{Journal of Mathematical Imaging and Vision}, vol.~51, no.~1, pp. 22--45, 2015.

\bibitem{corbin2023assessing}
A.~Corbin and O.~Marques, ``Assessing bias in skin lesion classifiers with contemporary deep learning and post-hoc explainability techniques,'' \emph{IEEE Access}, vol.~11, pp. 78\,339--78\,352, 2023.

\bibitem{suiccmez2023detection}
{\c{C}}.~Sui{\c{c}}mez, H.~T. Kahraman, A.~Sui{\c{c}}mez, C.~Y{\i}lmaz, and F.~Balc{\i}, ``Detection of melanoma with hybrid learning method by removing hair from dermoscopic images using image processing techniques and wavelet transform,'' \emph{Biomedical Signal Processing and Control}, vol.~84, p. 104729, 2023.

\bibitem{mirikharaji2023survey}
Z.~Mirikharaji, K.~Abhishek, A.~Bissoto, C.~Barata, S.~Avila, E.~Valle, M.~E. Celebi, and G.~Hamarneh, ``A survey on deep learning for skin lesion segmentation,'' \emph{Medical Image Analysis}, vol.~88, p. 102863, 2023.

\bibitem{tanveer2024comprehensive}
N.~Tanveer, N.~Tariq, A.~Akram, and F.~Shabbir, ``Comprehensive review on u-net architectures for skin lesion segmentation and its variants,'' \emph{Journal of Computing \& Biomedical Informatics}, vol.~7, no.~02, 2024.

\bibitem{yao2024cnn}
W.~Yao, J.~Bai, W.~Liao, Y.~Chen, M.~Liu, and Y.~Xie, ``From cnn to transformer: A review of medical image segmentation models,'' \emph{Journal of Imaging Informatics in Medicine}, vol.~37, no.~4, pp. 1529--1547, 2024.

\bibitem{huang2024comparative}
Y.~Huang, J.~Zou, L.~Meng, X.~Yue, Q.~Zhao, J.~Li, C.~Song, G.~Jimenez, S.~Li, and G.~Fu, ``Comparative analysis of imagenet pre-trained deep learning models and dinov2 in medical imaging classification,'' \emph{arXiv preprint arXiv:2402.07595}, 2024.

\bibitem{deng2009imagenet}
J.~Deng, W.~Dong, R.~Socher, L.-J. Li, K.~Li, and L.~Fei-Fei, ``Imagenet: A large-scale hierarchical image database,'' in \emph{2009 IEEE conference on computer vision and pattern recognition}.\hskip 1em plus 0.5em minus 0.4em\relax Ieee, 2009, pp. 248--255.

\bibitem{lin2014microsoft}
T.-Y. Lin, M.~Maire, S.~Belongie, J.~Hays, P.~Perona, D.~Ramanan, P.~Doll{\'a}r, and C.~L. Zitnick, ``Microsoft coco: Common objects in context,'' in \emph{European conference on computer vision}.\hskip 1em plus 0.5em minus 0.4em\relax Springer, 2014, pp. 740--755.

\bibitem{everingham2010pascal}
M.~Everingham, L.~Van~Gool, C.~K. Williams, J.~Winn, and A.~Zisserman, ``The pascal visual object classes (voc) challenge,'' \emph{International journal of computer vision}, vol.~88, no.~2, pp. 303--338, 2010.

\bibitem{alipour2024skin}
N.~Alipour, T.~Burke, and J.~Courtney, ``Skin type diversity in skin lesion datasets: A review,'' \emph{Current Dermatology Reports}, vol.~13, no.~3, pp. 198--210, 2024.

\bibitem{khanra2022survey}
S.~Khanra, M.~Kuila, S.~Patra, R.~Saha, and K.~G. Dhal, ``Survey on computational techniques for pigmented skin lesion segmentation,'' \emph{Optical Memory and Neural Networks}, vol.~31, no.~4, pp. 333--366, 2022.

\bibitem{zhou2024auto}
S.~Zhou, H.~Li, W.~Sun, F.~Zhou, and K.~Xiao, ``Auto-white balance algorithm of skin color based on asymmetric generative adversarial network,'' \emph{Color Research \& Application}, 2024.

\bibitem{mbatha2024skin}
S.~K. Mbatha, M.~J. Booysen, and R.~P. Theart, ``Skin tone estimation under diverse lighting conditions,'' \emph{Journal of Imaging}, vol.~10, no.~5, p. 109, 2024.

\end{thebibliography}

\vspace{-10mm}

\begin{IEEEbiographynophoto}{Kuniko Paxton} is a PhD candidate in Digital and Physical Sciences at the University of Hull, and the funding institution is DAIM. The research interests are Pruning Learning, Fairness in AI, Explainability, and Distributed Learning. Contact her at k.azuma-2021@hull.ac.uk
\end{IEEEbiographynophoto}

\begin{IEEEbiographynophoto}{Koorosh Aslansefat} is an assistant professor of computer science at the University of Hull, HU6 7RX Hull, U.K., Co-Leader of the Dependable Intelligent Systems research centre. His research interests span artificial intelligence safety, Markov modelling, and real-time dependability analysis. Aslansefat received his PhD in computer science from the University of Hull. He is a Senior Member of IEEE. Contact him at K.Aslansefat@hull.ac.uk
\end{IEEEbiographynophoto}

\begin{IEEEbiographynophoto}{Dhavalkumar Thakker} is a Professor of Artificial Intelligence (AI) and the Internet of Things (IoT) at the University of Hull, where he leads a group focused on Responsible Artificial Intelligence. His research emphasizes AI Explainability, AI Safety, and Fairness. With nearly two decades of experience, Dhavalkumar has been at the forefront of innovative solutions through funded projects. His interdisciplinary research spans Generative AI and the applications of Edge computing alongside IoT technologies.  He has a track record in leveraging AI for Social Good, notably in Smart Cities, Digital Health, and the Circular Economy. Contact him at D.Thakker@hull.ac.uk
\end{IEEEbiographynophoto}

\begin{IEEEbiographynophoto}{Yiannis Papadopoulos} is Professor of Computer Science and Leader of the Dependable Intelligent Systems (DEIS) Research Group at the University of Hull in the U.K. For over 30 years, Papadopoulos and his research group have pioneered cutting-edge model-based, bio-inspired and statistical technologies for the analysis and design of dependable engineering systems, with a recent intense focus and contributions towards achieving trustworthy, safe AI. Many of the software tools that they have crafted, including HiP-HOPS and EAST-ADL, have become commercial and are used in transport and other industries. Contact him at Y.I.Papadopoulos@hull.ac.uk
\end{IEEEbiographynophoto}

\begin{IEEEbiographynophoto}{Amila Akagic} is professor who received the B.Sc. (2006) and M.Sc. (2009) degrees in electrical engineering, with a focus on computer science and informatics, from the University of Sarajevo. She received the Ph.D. degree from Keio University, Japan, in 2013. She is currently a Full Professor in the Department of Computing and Informatics at the University of Sarajevo. Her research interests include artificial intelligence, machine learning, deep learning, computer vision, image processing, and digital signal processing.
\end{IEEEbiographynophoto}

\begin{IEEEbiographynophoto}{Medina Kapo} received her B.Sc. degree in 2021 and M.Sc. degree in 2023 in Computer Science and Informatics from the Faculty of Electrical Engineering, University of Sarajevo. She is currently employed as a Teaching Assistant at the Faculty of Electrical Engineering, University of Sarajevo, where she is also pursuing her Ph.D. studies in Computer Science and Informatics. Her research interests include computer vision, deep learning, machine learning, the application of artificial intelligence models and methods in medicine, and database systems.
\end{IEEEbiographynophoto}

\end{document}